\documentclass[manuscript, screen]{acmart}
\usepackage{svg}
\usepackage{graphicx} 

\usepackage{booktabs}  
\usepackage{array}     

\definecolor{DarkGreen}{rgb}{0.1,0.5,0.1}

\setcopyright{none}
\renewcommand\footnotetextcopyrightpermission[1]{}
\usepackage{tikz}
\usetikzlibrary{shapes.geometric, arrows.meta, positioning, fit, backgrounds}

\definecolor{cgray}{cmyk}{0,0,.5,.1}
\definecolor{agray}{cmyk}{.7,0,0,.2}

\title{Validity Is What You Need}
\author{Sebastian Benthall}
\affiliation{\institution{Monitaur}
\country{USA}}
\affiliation{\institution{International Computer Science Institute}
\city{Berkeley}
  \state{California}
  \country{USA}}
\affiliation{\institution{New York University School of Law}
\city{New York}
  \state{New York}
  \country{USA}}
\email{spb413@nyu.edu}

\author{Andrew Clark}
\affiliation{\institution{Monitaur}
\country{USA}}

\date{October 2025}

\begin{abstract}
    While AI agents have long been discussed and studied in computer science, today's Agentic AI systems
    are something new. We consider other definitions of Agentic AI and propose a new realist definition.
    Agentic AI is a software delivery mechanism, comparable to software as a service (SaaS), which puts
    an application to work autonomously in a complex enterprise setting.
    Recent advances in large language models (LLMs) as foundation models have driven excitement in Agentic AI.
    We note, however, that Agentic AI systems are primarily applications, not foundations, and so their success depends on validation by end users and principal stakeholders.
    The tools and techniques needed by the principal users to validate their applications are quite
    different from the tools and techniques used to evaluate foundation models.
    Ironically, with good validation measures in place, in many cases the foundation models can be replaced with much simpler, faster, and more interpretable models that handle core logic.
    When it comes to Agentic AI, validity is what you need.
    LLMs are one option that might achieve it.
\end{abstract}

\begin{document}

\maketitle

There is undeniably a current wave of excitement and investment in Agentic AI (Pounds, 2024), which is envisioned as an application of foundation models, such as LLMs, to perceive, reason, act, and learn. We examine several definitions of Agentic AI, as well as industry trends and arguments in the research literature, to assess the current state and future of Agentic AI. After considering definitions of Agentic AI from computer science, law, and industry, we propose our own ‘realist’ definition of agentic AI. Agentic AI, today, refers to a software delivery mechanism, similar to SaaS, which puts a complex, multi-step application to work within or between enterprises or their customers. It aspires to replace or significantly augment employees in a complex environment with multiple stakeholders and limits of information flow. Building an agentic AI in those conditions is much like the economic practice of mechanism design: building the mechanism that, when used, optimizes the design objectives. 

Given the power of AI systems to optimize, it has long been recognized that one of the greatest challenges of AI is alignment or, more technically, specifying the system’s goals \cite{ji2023ai}.
Agentic AI, being situated and applied in complex domains, requires a complex specification which must be validated and verified at the application level by the end-using person. Thus, the frontier for innovation in Agentic AI is in the design and validation process, turning social and organizational goals into metrics of success and operational tests. More complex agentic applications will require more complex testing and governance systems, and these testing and governance systems will be what trains the foundational technologies to serve as applications. We propose a multi-stage process for designing agentic systems, and question how AI can improve on this design and governance process.

The irony of agentic AI is that while its use is motivated by the new capacity of Foundational AI technologies, effective AI validation may well reduce the need for these foundation models over time. When goals are well specified, AI systems must use specific, rather than general, intelligence, and compete on performance guarantees and marginal efficiency. Smaller, interpretable models are likely to be the key components of Agentic AI systems that endure to maturity.

\section{Agency and AI}

This paper is written during the first point in history when “agentic AI” has entered the mainstream, industrial vocabulary. However, it is certainly not the first time artificially intelligent agents have been discussed in the AI scientific community. One challenge with reasoning about “Agentic AI” is that the terms “agent” and “intelligent” are evolving. We note that Larry’s Tesler’s adage that “AI is whatever hasn't been done yet” withstands the test of time – our understanding of “intelligence” often excludes those tasks that automation has mastered. Likewise, we consider how the meaning of artificial “agency” has evolved over time. We consider classical computer science, legal, contemporary computer science, and industry definitions. We then clarify how we will use the term, positing a realistic definition fit to the current business use cases.

The word “agent” comes originally from the Latin verb \emph{agere}, meaning ``to do or to drive'', and the past participle \emph{agens}, meaning ``doing'' or “one who is doing”. In this basic sense, anything that does anything is an agent. The simplest model of an artificial agent is based on a sensorimotor loop, with sensors providing information from the environment, some kind of decision process (which may be entirely reflexive) determining next actions, and actuators which translate those actions back into the environment. 

For decades in the AI community, an intelligent agent has been defined to be an agent which acts effectively and purposefully to achieve goals. The very widely used Russell and Norvig textbook \cite{russell1995artificial} uses the “intelligent agent” as its unifying theme across many paradigms of AI. They take care to distinguish agents from mere programs. Agents, unlike programs, operate under autonomous control, perceive their environment, persist for a prolonged time period, adapt to change, and take on goals. Rational agents are those that act to achieve the best expected outcome. In more recent work, \citet{russell2019human} emphasizes that agents are goal-oriented. Russell’s sense of agency is closely connected to the formal model of a utility maximizing agent, and especially one that maximizes utility over time. These themes are reiterated by other contemporary researchers (such as \citet{chan2023harms}) who distinguish “agency” from mere software. “Agents” accomplish goals less through specification of operations, and more through goal directedness. They also act with less human mediation, and more long-term planning.
This definition of intelligent agency fits well into the science of rational behavior that has been well established over time. The Bellman equation, developed by \citet{bellman1954theory}, forms the mathematical backbone of both economic optimization and modern reinforcement learning:
$$ V(x) = \max_{a \in \Gamma (x) } \{ F(x,a) + \beta V(T(x,a)) \} $$
Where $V(x)$ is the value function at state x, $F(x,a)$ is the immediate reward for taking action a in state x, $\beta$ is the discount factor, and $T$ is the transition function to next period states. In this paradigm of agent rationality, the agent is trained to learn an optimal policy that chooses an optimal action a at any state x.

It is no accident that “agentic AI” often builds on reinforcement learning today.
However, these criteria of ‘agency’ are not as precise as they may look. There are many trivial ways to satisfy a Bellman equation. Modern LLMs are models trained to minimize a loss function on the task of autoencoding massive datasets. A user of a modern chatbot queries these models and gets the “best” trained response, but this is a form of pattern matching \cite{shojaee2025illusion} rather than reasoning towards some other externally specified goal. Because these services are equipped with “like” and “unlike” buttons, or other ways of allowing users to flag their (dis)approval of responses, there is a widespread practice of Reinforcement Learning from Human Feedback (RLHF) \cite{bai2022training}, whereby these chatbots are gradually trained, much like social media curation algorithms, to maximize positive user engagement. They are already, by this definition, agents. However, the more recent definitions of Agentic AI differentiate between this model of AI interaction and others involving more autonomy and orchestration.

A related but different sense of “agency”, that comes from law and economics, is that of “a person employed to do or act on behalf of another”, where the other is called the principal. This sense of agent-for clarifies that the agent does not act to achieve their own goals, but some external  purpose that is meaningful to somebody else. This sense of agency has entered the scholarship on AI governance \cite{hadfield2019incomplete, hadfield2021principal, kolt2025governing}. In some legal contexts, such as fiduciary relations, agency obligations imply challenging requirements for AI agents \cite{benthall2023designing}. However, in almost all contexts the question of AI alignment, or validating that an AI is trying to accomplish what the user or principal wants it to be doing, is a challenging aspect of design and implementation \cite{leike2018scalable}.

As of the mid-2020’s, “agentic AI” has been promoted by industry and with it has come further refinements about what it means. OpenAI researchers \cite{shavit2023practices} have usefully defined agenticness as “the degree to which a system can adaptably achieve complex goals in complex environments with limited direct supervision.” This definition involves several dimensions of agenticness: goal and environment complexity, adaptability, and independent execution. These are all matters of degree, not matters of kind. This definition is motivated in part by how new LLM based agents represent advances along each of these dimensions. LLMs provide a high-dimensional and well-trained space of internal representations, which enables both goal and environment complexity. They can be retrained using finetuning data sets and human feedback – a form of adaptability. LLMs can be integrated with enterprise data and end users in novel ways using natural language interfaces, and also be equipped with “tools” which allow them to perform more computational functions besides generation or prediction – which are all elements of independent execution.

However, there are serious limitations facing LLMs when they are tasked with solving complex problems, such as the finite size of the context window \cite{shojaee2025illusion}.
Today (and for who knows how long), “agentic AI” often refers to the orchestration of multiple calls to an LLM, or even multiple distinct AI systems. Notable work in this area includes AutoGen \cite{wu2024autogen}, ChatDev \cite{qian2023chatdev}, and MetaGPT \cite{hong2023metagpt}. As an alternative to a monolithic LLM-based system that relies on a large context window for a complex query, decomposing a problem into specific subtasks and relaying them to dedicated “agents” is, in some respects, simply a matter of software architecture that is analogous to breaking a complex program into subroutines, or dividing a monolithic web service into “microprocesses” dedicated to discrete functions. While there are often benefits, with respect to performance or maintainability, to such decompositions, these do not translate directly into any qualitative difference in responsibility or entail a difference in motivation or goal-orientation. Nevertheless, these orchestrated systems are frequently called “multi-agent systems” today. This sense of “agency”, common in today’s industry lexicon, is far more prosaic and watered down than its use in other scientific literature.

Outside of academia, a different framing is threaded through industry sources. NVIDIA \cite{pounds2024agentic}, which stands to benefit from Agentic AI adoption and promotes it, defines Agentic AI as that which will:
\begin{itemize}
\item Perceive: Gather data from multiple inputs/stimuli
\item Reason: Make a reasoned inferences about the world
\item Act: Use API calls and integrations to execute tasks
\item Learn: Refine itself based on feedback
\end{itemize}

Whereas traditional AI models are often deployed to forecast future outcomes, classify documents, and generate text or images, “agentic AI” in this sense is marketed to replace or significantly augment employees, especially in an enterprise context. There are successful deployments of agentic LLMs for writing software documentation, providing engineering support, sorting through documents, and helping to process insurance claims, though corporate adoption has been off to a rocky start.

In this paper, we propose a \emph{realist} definition that is based on the emerging reality of business use cases. Contemporary “AI systems” are not theoretical constructs, they are real software systems operating through well-understood protocols. \emph{Agentic AI} refers, in practice, to a software service model, similar to “software as a service” (SaaS), where the AI product or service takes multiple steps of action in an embedded way in an enterprise’s business processes. By this definition, what distinguishes agentic AI is not especially the challenge of its environment or autonomy. An autonomous deep sea exploration robot seeking to collect samples of unknown aquatic life forms would  independently navigate and adapt to a complex environment, and be extremely cool. But it is not “agentic AI” in the business-relevant sense. “Agentic AI” faces complexity because of how it is situated \cite{suchman2007human} in real enterprise contexts, which are by nature multi-agent and sociotechnical. And it manages that complexity by taking multiple, connected actions in context.

\begin{table}[h]
\centering
\caption{Definitions of Intelligent Agents Across Different Perspectives}
\label{tab:agent-definitions}
\begin{tabular}{@{}p{0.15\linewidth}p{0.55\linewidth}p{0.22\linewidth}@{}}
\toprule
\textbf{Perspective} & \textbf{Definition} & \textbf{Sources} \\
\midrule
Classic & An intelligent agent is a system with a sensorimotor loop that acts to achieve specified goals. & \cite{russell1995artificial} \\
\addlinespace
Legal & An agent is a person who acts on behalf of a principal, performing in their best interests. & \cite{hadfield2021principal, kolt2025governing} \\
\addlinespace
Research & Agentic AI is adaptive, has limited supervision, achieves complex goals in a complex environment & \cite{shavit2023practices} \\
\addlinespace
Industry & Agentic AI perceives, reasons, acts, and learns, with an LLM & \cite{pounds2024agentic} \\
\addlinespace
Realist & Agentic AI is a service delivery mechanism for multi-step AI tools into a realistic enterprise setting. & \\
\bottomrule
\end{tabular}
\end{table}

\section{Foundations and Applications}

\begin{figure}
    \centering
    \includegraphics[trim={4cm 5cm 2cm 0cm},width=0.9\linewidth]{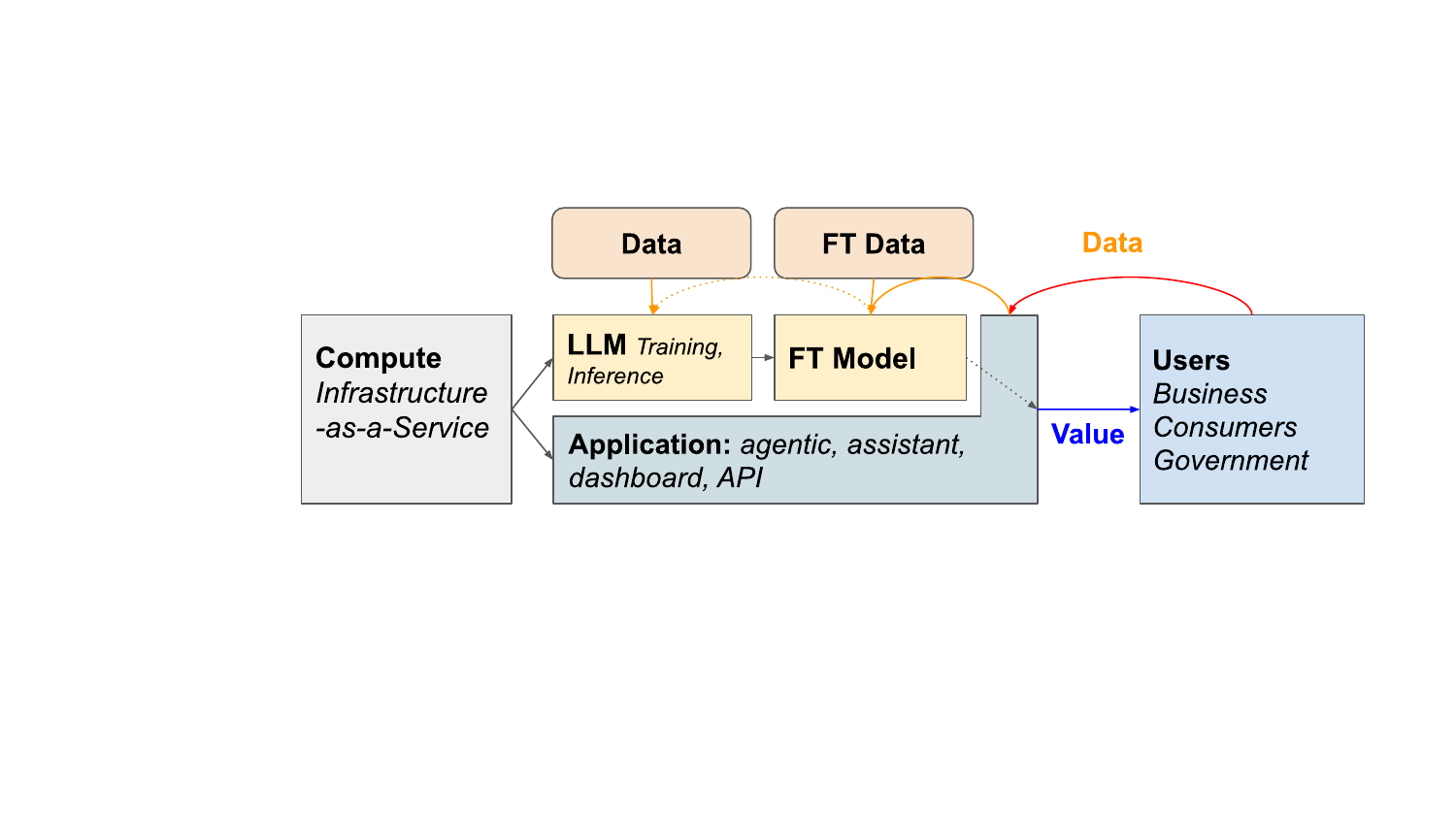}
    \caption{The supply chain of Agentic and other AI applications that use an underlying foundation model like an LLM. There are three sources of data for the AI application: pretraining data for the LLM, finetuning (FT) data for a finetuned model, and user data. What matters for the user is the value of the total application. The question facing the AI application designer is what information, and which data sources, are useful for delivering value in the enterprise context. Since dependence on a large foundation model can involve additional expenses (such as inference costs on large models, or licensing fees on training data, which are passed through to the downstream users), there is an incentive for the Agentic AI designer to minimize dependence on larger models.
}
    \label{fig:ai-supply-chain}
\end{figure}

If industry is able to deliver on the promise of Agentic AI, it will unlock tremendous value. That is because Agentic AI serves the specialized, specific, situated needs of stakeholders in their enterprise setting, and these stakeholder needs are ultimately the source of value.

It is somewhat ironic that this wave of excitement about Agentic AI”has arisen because of a very different form of technology: the foundation model \cite{bommasani2021opportunities}. Large Language Models (LLMS) and other foundation models are trained on vast data sets from every available context. Typically, they are trained to autoencode the data, meaning that they can predict the next token of the data given a sample or context \cite{vaswani2017attention, radford2018improving}. This training regime means that as much of the information in the training set is retained in the model as possible. It is then expected that the foundation model will be adapted to particular tasks.

In the simplest use of an LLMs, the model is prompted to generate digital content. Open source LLMs can be used directly in this way. But this is not how these systems are most typically used. Even consumer ready chat applications marketed through the web have layers of fine tuning, pretraining, and additional context which adapt the generated output to the technology and character of the application. These products and services aim to be of “general” use and suggest “general” intelligence. More advanced versions of these products act with more independence, in a more complex environment, because they are capable of searching the web and summarizing the results, or inspecting files. These services rely on the user to provide the prompts which allows it to deliver value.

The foundation model approach is motivated by economies of scale. Now notorious empirical scaling laws \cite{kaplan2020scaling} have led industry to pursue neural language models with larger and larger parameter sizes, training data sets, and computing power. There is also a professional tendency for computer science research more generally to focus on the abstract, formal, and portable, rather than the specifics of realistic context \cite{green2020algorithmic}.
These two tendencies – of computer science researchers towards abstraction, and of industry’s pursuit of economies of scale in technology – motivate new foundation models and platform services on which Agentic AI could, theoretically, be built. It is through Agentic AI that these models and services will find value. But the bridging of that last mile between these general services and the end user, whose needs are specific and situated, is the hardest and most difficult part of the value chain. These last parts of the value chain are differentiated by who has access to what information and whose incentives are at work in directing the design and use of the technology.

\section{Limits from Pretraining to Validation}

Because of the popularity and promise of LLMs, there is now a tremendous effort to demonstrate through benchmarking and evaluations their general applicability and reasoning power \cite{chollet2019measure, chollet2024arc, perlitz2024benchmark, shankar2024validates, white2024livebench}.
The case for this effort, which is to some extent borne out by empirical data, is that with a large and diverse enough training data set, the pretrained representations will be sufficient for any of a large variety of general tasks with only a small amount of careful finetuning \cite{zhou2023lima}, though has not yet borne out to be the case in many domains of interest \cite{huang2025crmarena, shojaee2025illusion}.

We question whether, in principle, any amount of demonstrated general performance for a large model can be sufficient for validating a specific enterprise end-user’s Agentic AI system. One argument for skepticism about Agentic AI validity comes from multi-step task complexity. We must distinguish between the core abilities of an LLM and its abilities when embedded in an agentic, multi-step context \cite{peng2024survey}. Even if there is a relatively high degree of accuracy on a single reasoning or inference task, these errors compound as the number of tasks in a multi-step process increases\footnote{As an intuition pump, if an agent has .9 accuracy on a single task, and performs with that accuracy in a sequence of four tasks, that can compound to .9 x .9 x .9 x .9 = .66 accuracy.}. Hence, enthusiasm about agentic AI efficacy must be checked by realistic thinking about task complexity.

Beyond that argument from task complexity, we see three substantial information theoretic challenges to valid Agentic AI which we claim require external validation structures to overcome.

\paragraph{1. Does the pretrained model have the information needed to accomplish the Agentic AI requirements?} LLMs are, due to their large parameterizations, able to hold large amounts of information from their training data. But this information is collected from general sources to create general representations. A specific Agentic AI deployment, on the other hand, will be deployed into a specific sociotechnical environment which we will presume is a proprietary context not available to the pretraining regime. Only the general, not the specific, aspects of the goals and environment will be available during pretraining. There will be an information gap. This gap is exacerbated by the reality of shift AI supply chains \cite{widder2023dislocated} and model obsolescence \cite{choksi2025brief}. Pretrained models will ratchet over time, and their performance can change in ways that disturb downstream uses. As LLMs become increasingly a form of platform with centralization and vendor lock-in, quality may deteriorate. In any case, performance is not guaranteed over time and requires continuous auditing.

\paragraph{2. Does the application designer have sufficient information about the pretrained model and the principal stakeholder needs?} As we have discussed, assuming that the pretrained model is sourced from a third-party and its subject to software updates, it is not guaranteed that its performance (let alone its accessibility) will be sustained over time. Hence, one of the application designer’s roles over the life cycle of the Agentic AI system will be verifying performance of the system according to automated tests, controls, and guardrails. These, along with any finetuning data sets for underlying LLMs, are all operational expressions of the principal stakeholder interests that are used for “alignment”. Critically, it is this technical material which encodes the stakeholder’s complex goals and environment, and the process for adapting and updating it over time, which determines the Agentic AI systems’ success.

\paragraph{3. Does the principal stakeholder have confidence in the Agentic AI system?} Ultimately, the principal stakeholder must have confidence in the Agentic AI system, and that cannot be supplied by the provider of the pretrained model because their guarantees are general, not specific to the specific use case. The principal stakeholder will have their own criteria or success and reliability which will be communicated to the application designer.

\paragraph{}We argue that much of the emphasis of Agentic AI development, which focuses on the general power of the pretrained foundation model, is misplaced. Rather, it is the processes and technologies that the application design uses to translate dynamic stakeholder needs into governance that are the crucial source of value in Agentic AI. These can include steps such as:
\begin{itemize}
\item Realistic scenario validation \cite{huang2025crmarena}
\item Ongoing Model Drift \cite{nelson2015evaluating, bayram2022concept}
\item Bias testing \cite{gallegos2024bias}
\item Guardrails \cite{kumar2025no}
\item Robustness to adversarial attacks \cite{shi2024large}
\end{itemize}

For a multi-step agentic program, the number of states of the system expands rapidly, at times requiring a state-based system to capture context when exacting standards are paramount \cite{shojaee2025illusion}.
For validation and ongoing monitoring, this significantly expands the surface area to manage. Non-standardized univariate model monitoring, as an example, needs to be performed at both an end to end as well as a discrete step level. End-to-end stress testing becomes of paramount importance, as well as “unit tests” at the ingestion and output for verifying critical information, such as credit card numbers, are correct, with “traditional” model monitoring methodologies required on the reasoning component.
Ultimately, end-to-end auditing principles from earlier machine learning and algorithmic implementations \cite{raji2020closing}  remain applicable to Agentic AI systems. A key difference is that Agentic AI is intended to operate within a dynamic, complex sociotechnical environment from the start, and has, in principle, the capacity to represent and reason about this context. This invites a large understanding of that context from the start of the application design.

\section{Agentic AI Application Design Process}

Given the above challenges to achieving Agentic AI from the perspective of the foundational or general technology provider, we instead offer a set of design heuristics to assist in agentic AI validation. We encourage the Agentic AI designer to see their task as one of “mechanism design”, as in economics and operations research. The intelligently designed application may or may not include a foundation model. It may or may not be subject to adaptation due to user feedback. What matters most is that it is an effective “agent” for the principal stakeholder, in an environment where multiple other stakeholders – including both humans and perhaps other “AI agents” – act to pursue their own goals. We urge the Agentic AI designer to:

\paragraph{1. Model the enterprise context as a multi-agent sociotechnical system.} Describe the 360 degree view of the operational context of the agentic AI.
\begin{itemize}
\item Who are the stakeholders, and how can they be categorized into different roles?
\item What are their incentives or goals?
\item Which stakeholders are the principals for whom the system is designed to benefit the most?
\item Who are the users of the system?
\item What resources are available to the system? Databases? The open Internet?
\item What are the mechanistic options and constraints in play?
\end{itemize}
The sourcing of this information is vital to the success of the Agentic AI system. [human centric design; participatory design; contextual integrity as sources here]

\paragraph{2. Define objectives in terms of that system and other constraints.} AI systems are trained to fulfil well-defined objective or utility functions. The model of the agentic AI’s context represents the universe of states of the world over which the principals have preferences. Enumerate the objectives of the principal, and operationalize each one in terms of the model, so that each is well-defined.

\paragraph{3. Check for feedback and leaks.} Assume that the Agentic AI is able to effectively pursue all of the stated objectives. Check to see if these results in the system exhibiting unexpected or perverse behavior.
\begin{itemize}
\item Will the system’s activity cause the distribution of data in its environment to shift away from its training data?
\item Will it bias its own data?
\item Will the behavior of stakeholders change as the result of new expectations set by the mechanism?
\end{itemize}
Operationalize these potential failures, and add these as additional guardrails, objectives, and constraints.

\paragraph{4. Build the agentic AI system.} Implement the system using the best available tools to meet the requirements.

\paragraph{5. Validate, verify, and train.} Observe the system in place, including the sensors and guardrails put in place. Ask stakeholders, especially principals, for their feedback. Are the assumptions of the system design correct? Is the system performing as expected? Can performance be improved by retraining?

\section{The irony: with strong validation, why the foundation?}

In the preceding sections, we have highlighted how successful agentic AI systems depend foremost on an understanding of real use cases, alignment with principal stakeholders, and an understanding of the governance surface area. We have outlined a procedure for designing a validation process that can be used to prepare and test the system for reliable value.

Notably, our procedure does not assume that the underlying computational engine of the AI system is an LLM or other foundation model. Natural language processing functions may be performed by smaller, more dedicated language models. Multi-step reasoning can be performed by more precisely designed expert systems that employ tools like dynamical systems, Markov Chains, dynamic programming, etc Indeed, many of the tools used in the validation system – those which operationalize the systems objectives and guardrails – can be used to provide consistent results.

These alternatives are important to consider given the many limitations and vulnerabilities of LLMs. LLMs can be “jailbroken” to perform behavior that is inconsistent with the task at hand \cite{wei2023jailbroken} and can hallucinate irrelevant and incorrect material \cite{zhang2025siren}. Reckless use of LLMs for web scraping can open a system up to security vulnerabilities due to prompt injection attacks. A holistic assessment of LLM capabilities in realistic professional settings found, among other things, “near zero confidentiality awareness” without targeted prompting \cite{huang2025crmarena}. In high-stakes settings, LLMs may not have the privacy or security guarantees needed to meet business requirements.

There is also evidence that LLMs may only provide the illusion of general “reasoning”, given how they are sharply limited by the size of their context windows \cite{shojaee2025illusion} For software coding assistance, where they are widely deployed and tested, they are still performing below grandmaster levels \cite{zheng2025livecodebench} and can have a negative impact on developer productivity \cite{becker2025measuring}. It remains to be seen if LLMs provide a viable foundation for agentic AI systems in real business use cases.

One alternative to LLMs that may efficiently address business use cases without some of the risks are Small Language Models (SLM) \cite{schick2020s}, which might be trained on more specialized data sets and have fewer parameters to begin with, or be ‘’quantized’’ versions of a larger model with lower weight resolution \cite{egashira2024exploiting}. It has been argued that these can be more appropriate for Agentic AI applications \cite{belcak2025small}. Performance can be achieved with models pretrained on high-quality, rather than fully general, training data \cite{gunasekar2023textbooks}. 

However, there are many other forms of AI that can provide the reasoning engine for such systems. These include utility theory, linear programming, dynamic programming, machine learning, and graph theoretic analysis. If the capabilities of foundation models succeed in exciting the demand for agentic AI, we anticipate that skillful application of these and other well-understood technologies will ultimately be responsible for satisfying it.

\section{Conclusion}

LLMs have created a new technical foundation for applications. This is creating a lot of activity in the application design space under the name of ‘agentic AI’. But this huge investment in finding applications of the new foundation is bound to run into the limits of application validation. The future of agentic AI is therefore in between requirements specification and expert systems to validate and verify situated applications that are assigned more complex tasks.  This is an open research area, akin to what’s currently done with ‘model evaluations’. Application evaluations which verify and validate are ultimately what any end user needs.

\bibliographystyle{plainnat}
\bibliography{validity}

\end{document}